\documentclass[conference]{IEEEtran}
\IEEEoverridecommandlockouts
\usepackage{cite}
\usepackage{amsmath,amssymb,amsfonts}
\usepackage{algorithmic}
\usepackage{graphicx}
\usepackage{textcomp}
\usepackage{xcolor}
\usepackage{url}
\usepackage{graphicx}
\usepackage[caption=false,font=footnotesize]{subfig}
\usepackage{booktabs}

\def\BibTeX{{\rm B\kern-.05em{\sc i\kern-.025em b}\kern-.08em
    T\kern-.1667em\lower.7ex\hbox{E}\kern-.125emX}}
\begin{document}

\title{A Bayesian Framework for Evaluating Scenario Compatibility in Generative Population Synthesis}

\author{
\thanks{The work was supported by the China Scholarship Council, China under Grant 202408320061 and the TRENoP and Digital Futures research centers at KTH Royal Institute of Technology, Sweden.} 

Zhenlin Qin\textsuperscript{1},
Leizhen Wang\textsuperscript{1,3},
Yancheng Ling\textsuperscript{1},
Zhenliang Ma\textsuperscript{*1,2}%
\thanks{\textsuperscript{1}Department of Civil and Architectural Engineering, KTH Royal Institute of Technology, Stockholm, Sweden.
(\{zhenlinq,yancheng,zhema\}@kth.se)}%
\thanks{\textsuperscript{2}Digital Futures, KTH Royal Institute of Technology, Stockholm, Sweden. (zhema@kth.se)}%
\thanks{\textsuperscript{3}Department of Data Science and Artificial Intelligence, Monash University, Melbourne, Australia.
(leizhen.wang@monash.edu)}%
\thanks{*Corresponding author.}  
\thanks{© 2026 IEEE. Personal use of this material is permitted.
Permission from IEEE must be obtained for all other uses, in any current
or future media, including reprinting/republishing this material for
advertising or promotional purposes, creating new collective works, for
resale or redistribution to servers or lists, or reuse of any copyrighted
component of this work in other works.}
}

\maketitle

\begin{abstract}
Scenario-based transportation analysis specifies future assumptions through aggregate population targets, whereas generative population synthesis models produce detailed individual-level realizations. When scenario targets are imposed on generative models, current practice relies on deterministic marginal calibration, implicitly assuming that the targets are compatible with the model’s learned structural support. However, whether scenario-level constraints lie within the generative support—and how strongly they distort structural uncertainty—remains largely unexamined. We propose an ensemble-based Bayesian updating framework to quantify scenario compatibility in conditional population synthesis. A population-aware conditional variational autoencoder is developed to learn a distribution over plausible population structures while preserving aggregate fidelity. An ensemble of realizations sampled from the learned prior provides an empirical approximation of structural uncertainty. Scenario targets are treated as probabilistic evidence over aggregate statistics, and posterior weights are obtained through Bayesian updating across the ensemble. Scenario compatibility is quantified using effective sample size (ESS), which measures posterior concentration and the compression of structural uncertainty induced by conditioning. Experiments demonstrate that scenario impact depends not only on target magnitude but also on alignment with the learned joint structure, and reveal structural failure modes when targets fall outside prior ensemble support. The proposed framework provides a probabilistic diagnostic model for evaluating scenario feasibility and structural consistency before downstream projection and transportation planning.

\end{abstract}


\section{Introduction}
Agent-based modeling (ABM) is a central paradigm in transportation research for analyzing individual-level behavior and emergent system dynamics under alternative planning and policy scenarios \cite{bonabeau2002agent,gilbert2005simulation,miller2018agent,wang2025agentic}. A fundamental prerequisite for ABMs is a realistic synthetic population that translates high-level scenario assumptions into detailed agent-level representations. In practice, transportation scenarios are typically expressed as aggregate targets, such as demographic composition, household structure, and mobility patterns. Operationalizing these targets requires population synthesis, which provides the concrete population realizations needed for subsequent simulation and policy evaluation.

Traditional population synthesis methods, including iterative proportional fitting (IPF), iterative proportional updating (IPU), and combinatorial optimization approaches, construct synthetic populations by deterministically matching aggregate constraints \cite{beckman1996creating,hermes2012review,muller2010population,barthelemy2013synthetic}. While these approaches effectively enforce marginal consistency, they typically produce a single calibrated population and do not explicitly represent the distribution of plausible population structures consistent with observed data. As a result, scenario targets are often imposed as hard constraints, without assessing whether they are probabilistically compatible with the underlying generative structure of the population model.

More recently, generative model-based approaches, such as variational autoencoders (VAEs) and generative adversarial networks (GANs), have been introduced to learn the joint distribution of population attributes from microdata \cite{aemmer2022generative,kim2023deep,garrido2020prediction,johnsen2022population}. By enabling stochastic sampling, these models naturally produce multiple population realizations, forming an ensemble of plausible synthetic populations. This generative perspective shifts population synthesis from producing a single calibrated solution to representing a distribution over possible population structures. However, existing approaches lack a principled framework to assess whether scenario-level assumptions are compatible with the generative support represented by the ensemble.

In this work, we interpret scenario targets as probabilistic evidence that updates the distribution over generated population realizations. Scenario information is incorporated through Bayesian updating across an ensemble of realizations, redistributing probability mass toward structures consistent with the targets. Under this formulation, compatibility is assessed through posterior concentration rather than deterministic marginal enforcement.

We implement this perspective within the SemaPop framework \cite{qin2026} using a variational backbone (SemaPop-VAE) and a population-aware objective that enhances aggregate structural fidelity, yielding SemaPop-VAE+. Given an ensemble sampled from the learned prior, scenario targets are formulated as likelihood functions over aggregate statistics, producing posterior weights across realizations. Compatibility is quantified using the effective sample size (ESS), which measures posterior concentration and uncertainty reduction under scenario conditioning.

The proposed framework serves as a probabilistic diagnostic layer that evaluates the structural compatibility of scenario assumptions with the learned generative support prior to downstream projection or policy simulation.

\section{Methodology}

\subsection{Problem Definition}

Let $X$ denote a complete population realization and 
$p_\theta(X)$ a learned generative distribution over plausible populations. 
Modern generative population models aim to approximate the joint distribution of demographic, household, and behavioral attributes from microdata. By enabling stochastic sampling, such models represent not a single calibrated population, but a distribution over many possible realizations.

In scenario-based transportation planning, particularly within ABM frameworks, future policy assumptions are typically specified as aggregate targets over population-level statistics, such as modal share, vehicle ownership rates, or demographic composition. 
Formally, we denote a scenario by $S = \{m_j\}_{j=1}^J$, where each $m_j$ is defined over population-level statistics $F_j(X)$. 

While generative models may reproduce observed marginal and joint dependencies, it remains unclear whether arbitrary combinations of such scenario targets are compatible with the distribution learned by the model. In other words, a scenario may correspond to population structures that are common under $p_\theta(X)$, or it may lie in regions that the model assigns very low probability.

This leads to the central question of this work:
\emph{Given a learned generative distribution $p_\theta(X)$, to what extent is a scenario $S$ supported by the distribution of plausible population realizations?}

Here, compatibility refers to how strongly a scenario reshapes the distribution over realizations. A compatible scenario can be satisfied by many plausible realizations and requires only moderate reweighting of probability mass. In contrast, a structurally tensioned scenario can be satisfied only by a small subset of realizations, concentrating probability mass on a narrow region of the distribution.

\begin{figure}[htbp]
\centering
\includegraphics[width=\columnwidth]{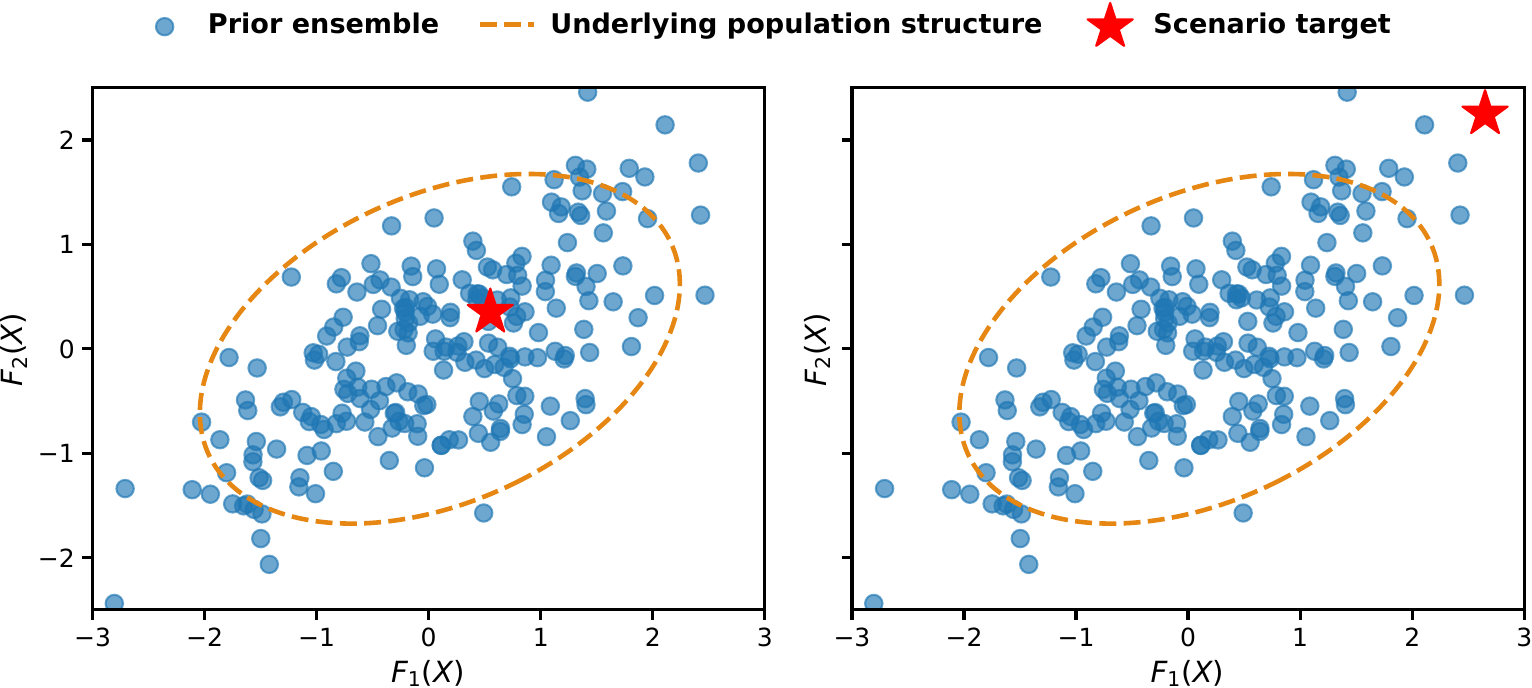}
\caption{Conceptual illustration of scenario compatibility in statistic space.
Left: compatible scenario aligned with learned support.
Right: structurally tensioned scenario outside learned support.}
\label{fig:scenario_compatibility}
\end{figure}

Fig.~\ref{fig:scenario_compatibility} provides a conceptual illustration in statistical space. The learned generative model defines a region of high-probability realizations.
A compatible scenario (left) lies within this region and aligns with a broad subset of realizations. A tensioned scenario (right) lies near or beyond its boundary, requiring substantial concentration on a limited feasible subset.

To address this problem, we adopt a probabilistic formulation in which scenario targets are treated as evidence that reshapes the distribution over realizations. This enables compatibility to be assessed by examining how the distribution changes under scenario conditioning, rather than by enforcing deterministic marginal alignment. 

To operationalize this formulation, we decompose the problem into two components: (i) learning a probabilistic representation of plausible population realizations, and (ii) quantifying how scenario targets reshape this distribution. Section B describes the generative synthesis model that approximates the prior distribution over realizations. Section C presents the compatibility measuring framework built upon this representation.

\subsection{Probabilistic Population Synthesis}

To enable compatibility-based uncertainty analysis, we require a generative model that represents a distribution over plausible population realizations. 
We denote by $p_\theta(X)$ the model-implied marginal distribution,
\begin{equation}
p_\theta(X) = \int p_\theta(X \mid z)\, p(z)\, dz,
\end{equation}
where $z$ denotes latent variables drawn from a predefined prior $p(z)$. 
The conditional distribution $p_\theta(X \mid z)$ is learned from observed microdata and captures joint structural dependencies among population attributes. 
Sampling from $p_\theta(X)$ yields an ensemble of plausible realizations approximating the learned generative support.

We introduce a variational instantiation of the SemaPop framework, termed SemaPop-VAE \cite{qin2026}. 
The model learns a joint latent representation of population attributes conditioned on semantic embeddings derived from persona descriptions. 
An encoder maps observed attributes to a latent space, and a decoder reconstructs demographic, household, and behavioral variables under latent regularization, enabling stochastic sampling from the learned distribution.

\begin{figure}[htbp] \centerline{\includegraphics[width=0.45\textwidth]{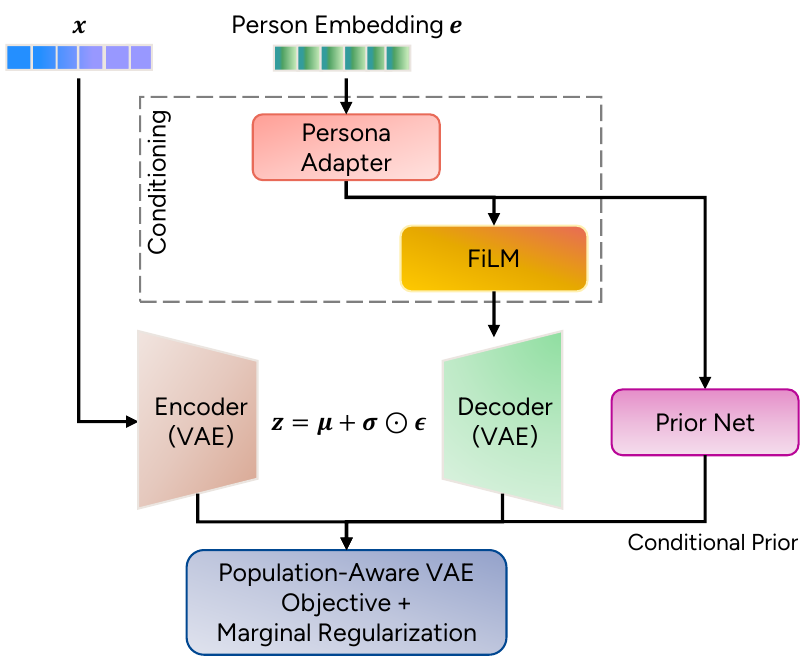}} \caption{Overview of the SemaPop-VAE+ model. The persona embeddings, semantic conditioning modules (persona adapter and FiLM), and conditional prior network follow the original SemaPop-VAE framework \cite{qin2026}. The key modification in this work is the population-aware variational objective with marginal regularization to enhance aggregate structural fidelity.} \label{vae_p_frame} \end{figure}

Standard variational objectives focus on per-sample reconstruction and may not sufficiently preserve aggregate structural properties. 
To enhance population-level fidelity, we introduce a hybrid reconstruction–likelihood objective combining mean-squared error (MSE) and Gaussian negative log-likelihood (NLL), forming a population-aware VAE objective. 
The resulting SemaPop-VAE+ model is illustrated in Fig.~\ref{vae_p_frame}.

Formally, for numeric variables we assume
\begin{equation}
p(\mathbf{x}^{\mathrm{num}} \mid \mathbf{z})
=
\mathcal{N}\!\left(
\boldsymbol{\mu}(\mathbf{z}),
\mathrm{diag}\!\left(\boldsymbol{\sigma}^2(\mathbf{z})\right)
\right),
\label{eq:gaussian_likelihood}
\end{equation}
with optional heteroscedastic variance parameterization.

\paragraph{Population-aware numeric objective}
We define
\begin{equation}
\mathcal{L}_{\mathrm{num}}
= \alpha_{\mathrm{num}}\Big(
\mathcal{L}^{\mathrm{num}}_{\mathrm{mse}}
+ \lambda_{\mathrm{nll}} \mathcal{L}^{\mathrm{num}}_{\mathrm{nll}}
\Big),
\label{eq:numeric_loss}
\end{equation}
where $\alpha_{\mathrm{num}}$ scales the loss, $\lambda_{\mathrm{nll}}$ balances likelihood regularization.



\paragraph{Overall training loss}
The full objective is
\begin{equation}
\mathcal{L}
= \mathcal{L}_{\mathrm{num}} 
+ \mathcal{L}_{\mathrm{cat}} 
+ \beta\,\mathcal{L}_{\mathrm{KL}}.
\end{equation}

Once trained, repeated sampling from $p_\theta(X)$ under fixed semantic conditions produces multiple complete population realizations. 
These realizations form an ensemble that approximates the learned distribution and defines the prior support for subsequent Bayesian updating under scenario constraints.

\subsection{Compatibility Measuring Framework}

Aggregate scenario targets are formulated as likelihood functions defined on population realizations sampled from the generative prior, enabling compatibility assessment in probabilistic rather than deterministic terms.

Let $\{\mathbf{X}^{(k)}\}_{k=1}^{K}$ denote an ensemble of realizations sampled from SemaPop-VAE+, forming an empirical approximation of the learned distribution $p_{\theta}(\mathbf{X})$ with equal prior mass $1/K$ assigned to each realization.

A scenario is specified by aggregate targets $\{m_j\}_{j=1}^{J}$ over population-level statistics. For each realization $\mathbf{X}^{(k)}$, the statistic $\hat{m}_j(\mathbf{X}^{(k)})$ is compared with the target through a Gaussian compatibility kernel:
\begin{equation}
p\!\left(S \mid \mathbf{X}^{(k)}\right)
=
\exp\!\left(
-\sum_{j=1}^{J}
\frac{\left(\hat{m}_j(\mathbf{X}^{(k)}) - m_j\right)^2}
{2\sigma_j^2}
\right),
\label{eq:scenario_likelihood}
\end{equation}
where $\sigma_j$ controls the tolerance around $m_j$. Realizations closer to the targets therefore receive higher likelihood. 

This formulation corresponds to the exponent of a diagonal multivariate Gaussian kernel in the space of aggregate statistics. 
When $J=1$, it reduces to a univariate Gaussian kernel; for $J>1$, it defines a diagonal multivariate form over multiple scenario targets, allowing simultaneous conditioning on several aggregate assumptions. 
It serves as a smooth compatibility function that quantifies the weighted squared distance between a generated population realization and a specified target scenario. 
It does not assume that realization-level aggregate statistics follow a joint Gaussian distribution centered at the targets.

The ensemble defines an empirical prior
\begin{equation}
\hat p(\mathbf{X})
=
\frac{1}{K}
\sum_{k=1}^{K}
\delta(\mathbf{X}-\mathbf{X}^{(k)}),
\label{eq:empirical_prior}
\end{equation}
where $\delta(\cdot)$ denotes a Dirac mass concentrated at each realization. 
This empirical measure converges to the generative distribution as $K$ increases.

Bayesian updating yields
\begin{equation}
p(\mathbf{X}\mid S)
\propto
p(S\mid \mathbf{X})\, p(\mathbf{X}).
\end{equation}

Substituting the empirical prior in Eq.~(\ref{eq:empirical_prior}) gives a posterior measure supported on $\{\mathbf{X}^{(k)}\}_{k=1}^K$,
\begin{equation}
p(\mathbf{X}\mid S)
\propto
\sum_{k=1}^{K}
\frac{1}{K}
p(S\mid \mathbf{X}^{(k)})
\delta(\mathbf{X}-\mathbf{X}^{(k)}).
\end{equation}

Therefore, the posterior mass assigned to each realization satisfies
\begin{equation}
p(\mathbf{X}^{(k)} \mid S)
\propto
p(S \mid \mathbf{X}^{(k)}) \cdot \frac{1}{K}.
\end{equation}

Since $1/K$ is identical for all realizations, it cancels under normalization. The posterior weights are therefore
\begin{equation}
\pi_k
=
\frac{
p\!\left(S \mid \mathbf{X}^{(k)}\right)
}{
\sum_{i=1}^{K}
p\!\left(S \mid \mathbf{X}^{(i)}\right)
},
\label{eq:posterior_weights}
\end{equation}
with $\sum_{k=1}^{K} \pi_k = 1$.

The posterior distribution is approximated by the weighted empirical measure
\begin{equation}
p(\mathbf{X} \mid S)
\approx
\sum_{k=1}^{K}
\pi_k \, \delta(\mathbf{X} - \mathbf{X}^{(k)}),
\label{eq:posterior_empirical}
\end{equation}
which preserves generative support while redistributing probability mass across realizations.

To quantify posterior concentration, we employ the effective sample size (ESS):
\begin{equation}
\mathrm{ESS}
=
\frac{1}{\sum_{k=1}^{K} \pi_k^2}.
\label{eq:ess}
\end{equation}

When $\pi_k \approx 1/K$, $\mathrm{ESS}\approx K$, indicating weak selective filtering and broad structural compatibility. 
Lower ESS values indicate that posterior mass is concentrated on a smaller effective subset of realizations, reflecting greater structural tension with the prior ensemble. 
Therefore, ESS provides an interpretable world-level measure of posterior concentration under scenario conditioning.

\begin{figure}[htbp]
\centering
\includegraphics[width=\linewidth]{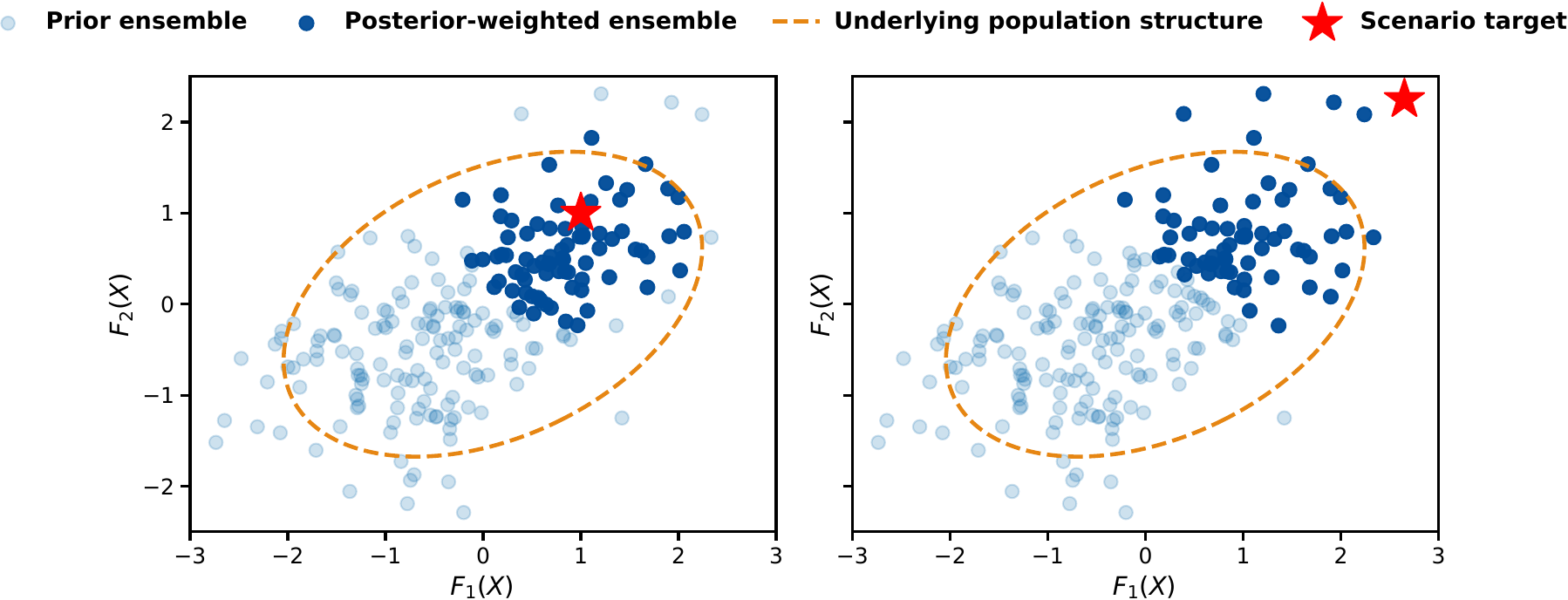}
\caption{Posterior concentration under scenario conditioning in statistic space. 
Light blue points denote the prior ensemble, and dark blue points indicate posterior-weighted realizations after Bayesian updating. 
In the compatible case (left), probability mass concentrates around the scenario target. 
In the structurally tensioned case (right), posterior ensemble is limited by the learned support, with only a small subset approaching the target.}
\label{fig:posterior_concentration}
\end{figure}

As illustrated in Fig.~\ref{fig:posterior_concentration}, Bayesian reweighting reallocates probability mass across the prior ensemble (light blue) toward realizations consistent with the scenario target. 
When the scenario is compatible (left), the target lies well within the learned structural support, and posterior adjustment induces only mild redistribution of weights. Most realizations retain comparable contributions, resulting in an ESS close to $K$. 
Under structural tension (right), however, the target approaches or exceeds the boundary of the learned support. Reweighting then becomes more selective, concentrating probability mass on a smaller effective subset of realizations and producing an ESS substantially smaller than $K$. 
Rather than enforcing deterministic aggregate matching through agent-level calibration, ESS characterizes how strongly scenario assumptions compress the ensemble within the model’s learned structural support.

\section{Case Study}
\subsection{Experiment Settings}
\paragraph{Data description}
All experiments are performed on a nationwide synthetic population dataset of Sweden, which is publicly released through Mendeley Data.\footnote{\url{https://data.mendeley.com/datasets/9n29p7rmn5/2}} 
The dataset contains a microscopically simulated population of more than 10 million individuals (agents) and is specifically constructed for activity-based transport modeling and large-scale population simulation analyses. 
The synthetic agents are calibrated to reproduce official Swedish demographic and travel statistics for 2018, thereby aligning the micro-level representations with observed aggregate population and mobility structures \cite{tozluouglu2023synthetic}.

The data are organized as a relational database comprising three linked tables—\emph{Person}, \emph{Household}, and \emph{Activity-travel}—which encode individual socio-demographic attributes, household context, and daily activity–travel behavior, respectively. A comprehensive description of the population attributes, including their semantic groupings and data types, is documented in our previous work \cite{qin2026}.

To construct the experimental dataset, we perform municipality-level stratified subsampling. Sweden consists of 290 municipalities, from each of which a fixed proportion (20\%) of agents is randomly drawn from the synthetic population. This proportional strategy preserves spatial heterogeneity and relative municipality sizes while reducing computational cost. The resulting sample is split into training (50{,}873), validation (20{,}262), and test (2{,}040{,}650) sets. The intentionally imbalanced split reflects a realistic population synthesis setting, where models are calibrated on limited microdata and generalized to a substantially larger target population.

\paragraph{Baselines}
We compare the proposed framework with representative probabilistic population synthesis methods spanning both classical and deep generative paradigms. All models are evaluated under a unified experimental protocol to ensure a fair comparison of their distributional fidelity and population synthesis performance.
\begin{itemize}
\item \textbf{TVAE} \cite{xu2019modeling}. 
A variational autoencoder designed for tabular data generation, which models the joint distribution through a latent-variable formulation optimized via reconstruction loss and KL-divergence regularization.

\item \textbf{BN} \cite{sun2015bayesian}. 
A Bayesian network–based population synthesis approach that factorizes the joint distribution into a directed acyclic graph of conditional dependencies, providing an interpretable probabilistic representation of individual and household attributes.

\item \textbf{BN-Copula} \cite{jutras2024copula}. 
A copula-enhanced Bayesian network model that separates marginal distributions from dependency structures, enabling flexible modeling of continuous-variable correlations and improved transferability under varying marginal constraints.

\item \textbf{SemaPop-VAE} \cite{qin2026}. 
An instantiation of the SemaPop framework employing a prior-conditioned VAE as the generative backbone. The training objective follows the TVAE formulation, reconstructing continuous attributes via Gaussian negative log-likelihood (NLL) in the standardized feature space.
\end{itemize}

\paragraph{Evaluation metrics}

We evaluate population synthesis performance using complementary metrics of distributional fidelity and structural validity. Distributional alignment between generated and reference populations (held-out test set) is quantified by the Standardized Root Mean Square Error at marginal (SRMSE-M) and bivariate-marginal (SRMSE-B) levels \cite{garrido2020prediction}. Structural feasibility and diversity are assessed using Precision, Recall, and F1 scores \cite{kim2023deep}. Implementation details follow \cite{qin2026}.

\paragraph{Implemetation details}
All experiments are conducted on a workstation equipped with four NVIDIA RTX 6000 Ada GPUs, providing a total of 192~GB of GPU memory. Persona representations are generated using the lightweight state-of-the-art large language model Qwen3-8B. During persona generation, we adopt a temperature of 0.9 and a top-$p$ value of 0.9, with a maximum of 512 newly generated tokens and the reasoning mode disabled. The detailed hyperparameter settings of SemaPop-VAE+ are summarized in Table~\ref{hyper_vae_p}.

\begin{table}[htbp]
\centering
\caption{Hyperparameters of SemaPop-VAE+.}
\label{hyper_vae_p}
\resizebox{0.57\linewidth}{!}{%
\begin{tabular}{ll}
\hline
Parameter             & Value           \\ \hline
Adapter hidden dim    & 1024            \\
Adapter condition dim & 128             \\
Adapter dropout       & 0.1             \\
FiLM feature dim      & (512, 512, 512) \\
PriorNet hidden dim   & 512             \\
Encoder hidden dim    & (512, 512, 512) \\
Decoder hidden dim    & (512, 512, 512) \\
Encoder dropout       & 0.1             \\
Decoder dropout       & 0.1             \\
$\beta$                  & 1.0               \\
$\lambda_m$             & 2.0               \\
$\alpha_{\mathrm{num}}$             & 120.0               \\
$\lambda_{\mathrm{nll}}$             & 0.001               \\
$\lambda_{\sigma}$             & 0               \\
Latent dim            & 128             \\
Learning rate         & 0.0002        \\
Batch size            & 512             \\
Epochs                & 300             \\
Optimizer             & Adam \\ \hline
\end{tabular}}
\end{table}

\begin{table}[htbp]
\centering
\caption{Generative performance comparison between the SemaPop-VAE+ and baseline models.}
\label{tab_perfcomp}
\resizebox{0.95\linewidth}{!}{%
\begin{tabular}{lccccc}
\hline
Model        & SRMSE\_M        & SRMASE\_B       & Precision      & Recall         & F1             \\ \hline
BN           & 0.0119          & 0.0679          & 4.57           & 79.60          & 8.64           \\
BN-Copula    & 0.0119          & 0.0679          & 4.66           & 79.26          & 8.81           \\
TVAE         & 0.0119          & 0.0672          & 6.74           & 79.32          & 12.43          \\
SemaPop-VAE  & 0.0114          & 0.0628          & 15.20          & 86.10          & 25.84          \\
SemaPop-VAE+ & \textbf{0.0097} & \textbf{0.0507} & \textbf{73.41} & \textbf{95.55} & \textbf{83.03} \\ \hline
\end{tabular}
}
\end{table}

\subsection{Generative Performance Comparison}

Table~\ref{tab_perfcomp} shows a consistent improvement in structural fidelity across model families. Classical graphical models (BN, BN-Copula) reproduce marginal statistics but struggle to capture complex joint distributions, resulting in weak structural feasibility. TVAE improves joint consistency but still produces many structurally invalid combinations, as reflected by its low precision and F1.

SemaPop-VAE further strengthens joint consistency under prior conditioning, while SemaPop-VAE+ substantially improves structural validity. It achieves the lowest marginal and joint discrepancies and dramatically higher precision and F1, indicating that the population-aware objective enables balanced reproduction of valid attribute combinations without sacrificing diversity (Recall).

\subsection{Scenario-Conditioned Population Ensemble Analysis}
To isolate posterior dynamics from potential training-induced biases, we construct a fixed evaluation pool of 10{,}054 agents from the held-out test set with aligned persona embeddings. Using only unseen individuals ensures that ensemble behavior reflects the learned generative distribution rather than memorization effects.

From this pool, we generate $K=100$ stochastic population realizations using SemaPop-VAE+. For each realization, a subsample of $N_{\text{sub}}=5000$ individuals is randomly drawn without replacement, and their attributes are synthesized conditional on the corresponding persona embeddings. The resulting ensemble forms an empirical distribution prior to scenario reweighting.

To characterize ensemble-level variability, we compute four realization-level summary statistics derived from model-generated attributes:
(i) PT activation rate, defined as the proportion of individuals with $Trips\_of\_PublicTransport \geq 1$; (ii) car-household activation rate, defined as the proportion of households with  $Number\_of\_cars\_of\_household \geq 1$;
(iii) mean age from $Age$; and
(iv) mean household size from $Household\_Size$.

For each realization, these statistics yield one value per metric. 
Under uniform weighting, the $K$ values form an ensemble distribution whose dispersion is summarized in Table~\ref{tab_disp}.

To examine the impact of scenario constraints, we introduce two shift scenarios in addition to the baseline (prior) case with uniform realization weights. Each scenario specifies absolute changes in public transport (PT) and car-household (CarHH) activation rates relative to their baseline means. 
Scenario~S1 applies $\Delta_{\mathrm{PT}} = +0.03$ and $\Delta_{\mathrm{Car}} = -0.03$, whereas Scenario~S2 applies $\Delta_{\mathrm{PT}} = +0.03$ and $\Delta_{\mathrm{Car}} = +0.03$. Posterior realization weights are computed using Gaussian likelihood functions centered at the shifted target means with standard deviation $\sigma = 0.015$ for each statistic. 
The complete specifications are summarized in Table~\ref{tab:scenario_settings}. The standard deviation $\sigma = 0.015$ is selected to balance scenario strictness and posterior diversity, providing sufficient discrimination across realizations without inducing weight degeneracy.

\begin{table}[htbp]
\centering
\caption{Prior ensemble variability of realization-level summary statistics ($K=100$). 
}
\label{tab_disp}
\resizebox{0.95\linewidth}{!}{%
\begin{tabular}{lcccc}
\hline
Summary Statistic             & Mean & Std    & Min    & Max    \\ \hline
PT Activation Rate            & 0.1909                     & 0.0037 & 0.1790 & 0.1992 \\
Car-Household Activation Rate & 0.4872                     & 0.0056 & 0.4720 & 0.5022 \\
Mean Age                      & 40.79                      & 0.2059 & 40.28  & 41.35  \\
Mean   Household Size         & 4.2034                     & 0.0469 & 4.0856 & 4.3104 \\ \hline
\end{tabular}
}
\end{table}

\begin{table}[htbp]
\centering
\caption{Scenario specifications for posterior reweighting. 
Shifts are relative to baseline means.}
\label{tab:scenario_settings}
\begin{tabular}{lccc}
\hline
Scenario & $\Delta$PT & $\Delta$CarHH & $\sigma$ \\
\hline
S0: Baseline & 0 & 0 & -- \\
S1: PT$\uparrow$, CarHH$\downarrow$ & +0.03 & $-0.03$ & 0.015 \\
S2: PT$\uparrow$, CarHH$\uparrow$   & +0.03 & +0.03 & 0.015 \\
\hline
\end{tabular}
\end{table}

\begin{table}[htbp]
\centering
\caption{Posterior statistics (mean) and effective sample size (ESS) across scenarios.}
\label{tab:scenario_results}
\begin{tabular}{lccc}
\hline
Statistic & Baseline & S1 & S2 \\
\hline
\multicolumn{4}{l}{\textit{Target statistics}} \\
PT activation rate    & 0.1909 & 0.1933 & 0.1920 \\
CarHH activation rate & 0.4872 & 0.4826 & 0.4905 \\

\hline
\multicolumn{4}{l}{\textit{Non-target statistics}} \\
Age                   & 40.79  & 40.80  & 40.78 \\
Household size        & 4.2034 & 4.1943 & 4.214 \\

\hline
\multicolumn{4}{l}{\textit{Posterior concentration}} \\
ESS                   & 100    & 46.8   & 59.7 \\
\hline
\end{tabular}
\end{table}

\begin{figure}[htbp]
\centering
\subfloat[S1: PT $\uparrow$ 0.03, CarHH $\downarrow$ 0.03]{
    \includegraphics[width=0.48\textwidth]{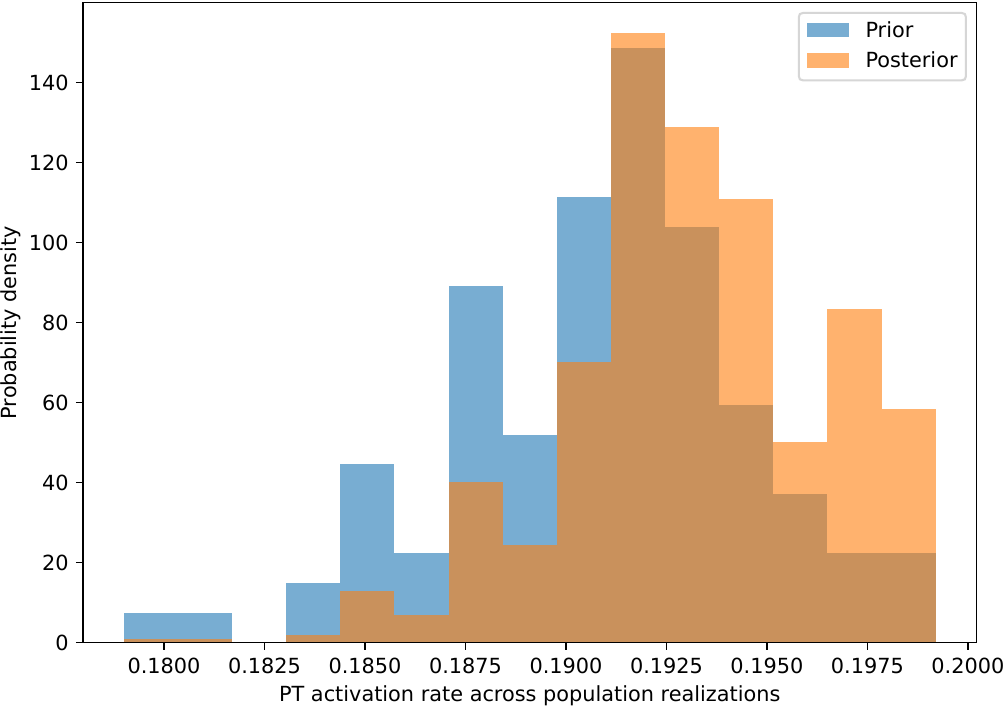}
    \label{fig:s1_pt}
}
\hfill
\subfloat[S2: PT $\uparrow$ 0.03, CarHH $\uparrow$ 0.03]{
    \includegraphics[width=0.48\textwidth]{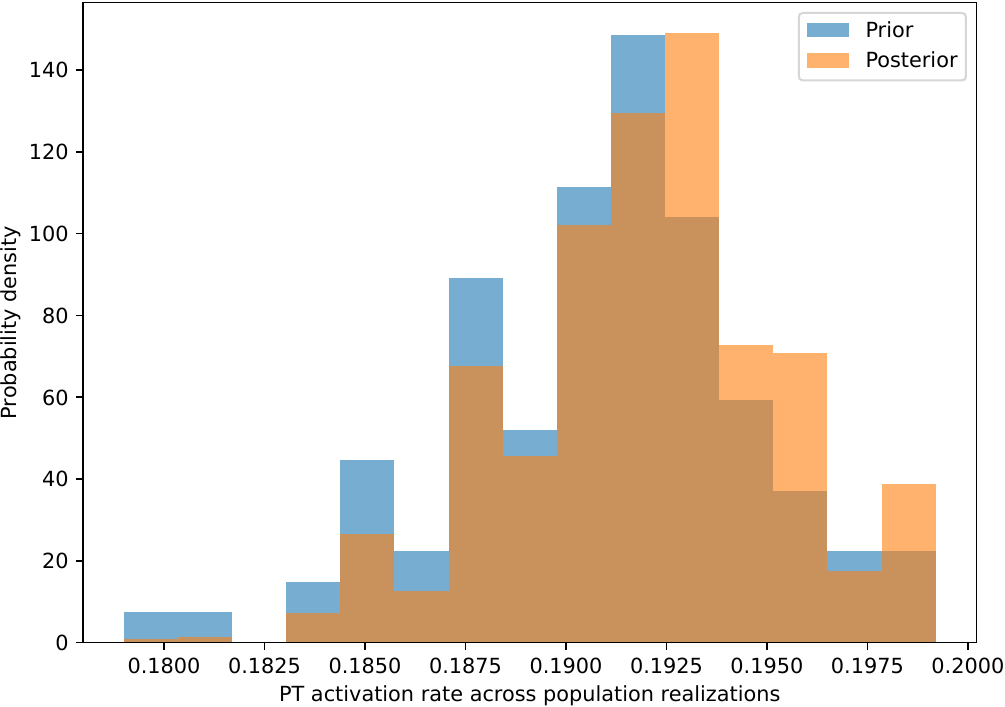}
    \label{fig:s2_pt}
}
\caption{Prior and posterior distributions of PT activation rates across population realizations under different scenario constraints. 
Posterior weights are computed using Gaussian likelihoods with $\sigma=0.015$.}
\label{fig:pt_prior_post}
\end{figure}

Table~\ref{tab:scenario_results} and Fig.~\ref{fig:pt_prior_post} illustrate the impact of scenario-based posterior reweighting. 
In both S1 and S2, posterior means shift in the intended directions, confirming that the likelihood reallocates probability mass toward scenario-compatible realizations, while $\sigma = 0.015$ maintains soft conditioning rather than exact target enforcement.

Differences in posterior concentration are evident from both ESS and distributional shape. 
S1 yields a lower ESS (46.8) than S2 (59.7) and exhibits a stronger right-tail concentration, indicating that only a narrower subset of realizations with relatively high PT activation rates can support the substitution constraint (PT$\uparrow$, CarHH$\downarrow$). 
By contrast, S2 remains compatible with a broader region of the ensemble. 
This difference can be interpreted as reflecting distinct mobility regimes: S2 may correspond to an overall expansion of mobility demand, whereas S1 requires tighter structural coupling between reduced car dependence and increased PT usage. 

The stronger concentration under S1 suggests a deeper structural shift in the population distribution. Only a narrow region of the learned support accommodates the S1 scenario targets, revealing structural tension with the prior. Further tightening of the targets may exceed this support and necessitate model retraining to capture alternative structural regimes.

\section{Conclusion}
This paper proposed an ensemble-based Bayesian updating framework to evaluate the compatibility between scenario-level targets and generative population structures. 
By treating scenario assumptions as probabilistic evidence over aggregate statistics, the approach enables soft posterior reweighting rather than deterministic marginal enforcement. 
The effective sample size
quantifies how scenario conditioning compresses structural uncertainty across realizations. 
Results show that scenario-induced shifts depend not only on target magnitude but also on alignment with the learned joint structure. In our experiments, the joint PT-increase and car-household reduction scenario requires more concentrated posterior support.

The framework provides a probabilistic diagnostic model for assessing the structural compatibility of scenario targets with the learned generative support before downstream population projection and transportation planning.



\bibliographystyle{IEEEtran}
\bibliography{_mybib}

\end{document}